\RequirePackage{etoolbox}
\csdef{input@path}%
{%
 {sty/},
 {img/},
}
\documentclass{elsevierbook}

\usepackage{biblatex}
\usepackage{soul}
\usepackage{dirtree}
\usepackage{multirow}
\usepackage{xcolor}
\usepackage{algorithm}
\usepackage{algpseudocode}

\begin{document}
%
\Mainmatter
  \begin{frontmatter}

\chapter*{Meta-Learning of NAS for Few-shot Learning in Medical Image Applications}\label{chap1}
\begin{aug}
\author[addressrefs={ad1}]%
  {\fnm{Viet-Khoa}   \snm{Vo-Ho}}%
\author[addressrefs={ad1}]%
  {\fnm{Kashu}   \snm{Yamazaki}}%
\author[addressrefs={ad2}]%
  {\fnm{Hieu}   \snm{Hoang}}%
\author[addressrefs={ad2}]%
  {\fnm{Minh-Triet}   \snm{Tran}}%
  \author[addressrefs={ad1}]%
  {\fnm{Ngan}   \snm{Le}}%

\address[id=ad1]%
  {Department of Computer Science \& Computer Engineering, University of Arkansas, Fayetteville, USA 72703 }%
\address[id=ad2]%
  {Department of Computer Science, VNUHCM-University of Science, Vietnam, }%
\end{aug}


\begin{abstract}
Deep learning methods have been successful in solving tasks in machine learning and have made breakthroughs in many sectors owing to their ability to automatically extract features from unstructured data. However, their performance relies on manual trial-and-error processes for selecting an appropriate network architecture, hyperparameters for training, and pre-/post-procedures. Even though it has been shown that network architecture plays a critical role in learning feature representation feature from data and the final performance, searching for the best network architecture is computationally intensive and heavily relies on researchers' experience. Automated machine learning (AutoML) and its advanced techniques i.e. \textit{Neural Architecture Search (NAS)} have been promoted to address those limitations. Not only in general computer vision tasks, but NAS has also motivated various applications in multiple areas including medical imaging. In medical imaging, NAS has significant progress in improving the accuracy of image classification, segmentation, reconstruction, and more. However, NAS requires the availability of large annotated data, considerable computation resources, and pre-defined tasks. To address such limitations, meta-learning has been adopted in the scenarios of few-shot learning and multiple tasks. 

In this book chapter, we first present a brief review of NAS by discussing well-known approaches in search space, search strategy, and evaluation strategy. We then introduce various NAS approaches in medical imaging with different applications such as classification, segmentation, detection, reconstruction, etc. Meta-learning in NAS for few-shot learning and multiple tasks is then explained. Finally, we describe several open problems in NAS.

\end{abstract}

\begin{keywords}
\kwd{Neural Architecture Search}
\kwd{NAS}
\kwd{Medical Imaging}
\kwd{Applications}
\kwd{AutoML}
\kwd{Meta Learning}
\end{keywords}

\end{frontmatter}

\section{Neural Architecture Search: Background}\label{sec1}
From the earliest LeNet \cite{lenet5} to the recent deep learning networks, designing network architecture heavily relies on prior knowledge and the experience of researchers. Furthermore, searching for optimal and effective network architecture is also time-consuming and computationally intensive due to the immersive amount of experiments for every architecture.
Automated machine learning (AutoML) is recently proposed to align with such demands to automatically design the network architecture instead of relying on human experiences and repeated manual tuning. Neural Architecture Search (NAS) is an instance of hyperparameter optimization that aims to search the optimal network architecture for a given task automatically, instead of handcrafting the building blocks or layers of the model \cite{cheng2020hierarchical}.

\begin{figure}[!t]
\begin{center}
  \includegraphics[width=0.8\linewidth]{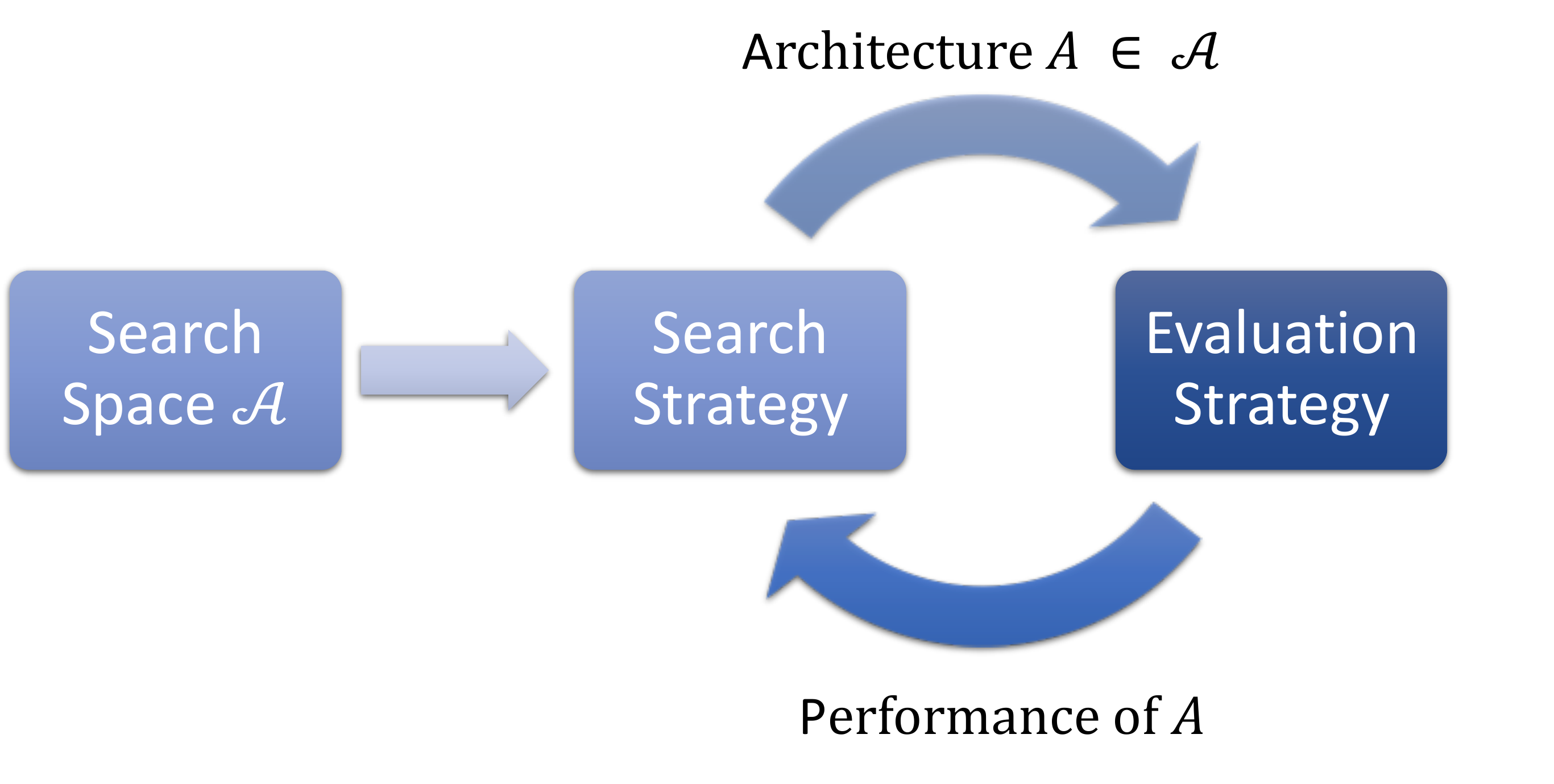}
\end{center}
  \caption{NAS algorithm with three components: search space, search strategy, and evolution strategy.}
\label{fig:NAS}
\end{figure}

NAS methods has already identified more efficient network architectures in general computer vision tasks. MetaQNN \cite{baker2016designing} and NAS-RL \cite{zoph2016neural} are the two earliest works in in the field of NAS. By automating the design of a neural network for the task at hand, NAS has tremendous potential to to surpass the human design of deep networks for both visual recognition and natural language processing \cite{liu2018progressive, chen2018searching, perez2018efficient, tan2019mnasnet}. This motivated various NAS applications in medical image tasks e.g. classification, segmentation, reconstruction. In a standard problem setup, the outer optimization stage searches for architectures with good validation performance while the inner optimization stage trains networks with the specified architecture. However, a full evaluation of the inner loop is expensive since it requires a many-shot neural network to be trained. 
A typical NAS technique contains two stages: the searching stage, which aims to find a good architecture, and the evaluating stage, where
the best architecture is trained from scratch and validated on the test data. Corresponding to two stages, there are three primary components: search space $\mathcal{A}$, search strategy, and evaluation strategy. Denote $A$ is single network in $\mathcal{A}$ i.e. $A \in \mathcal{A}$. Let $S^{train}$ and $S^{val}$ are two datasets corresponding to the training and testing sets. The performance of the network $A$, which is trained on $S^{train}$ and evaluated on $S^{val}$, is measured by function $\mathcal{F}$. In general, NAS can be mathematically formulated as follows:
\begin{equation}
    \underset{A}{\operatorname{argmax}} = \mathcal{F}(S^{val}(S^{train} (A)))
\end{equation}
Fig.\ref{fig:NAS} visualizes a general NAS architecture with three components i.e. search space, search strategy, and evaluation strategy.

\subsection{Search Space}
In principle, the search space specifies a set of operations (e.g. convolution, normalization, activation, etc.) and how they are connected as well as it defines which architectures can be represented. Thus, search space design has a key impact on the final performance of the NAS algorithm. In general, the search space defines a set of configurations that can be continuous or discrete hyperparameters \cite{liu2018darts} in a structured or unstructured fashion. In NAS, search spaces usually involve discrete hyperparameters with a structure that can be captured with a directed acyclic graph (DAG) \cite{pham2018efficient, liu2018darts}. The most naive approach to designing the search space for neural network architectures is to depict network topologies, either CNN or RNN, with a list of sequential layer-wise operations, as can be seen in \cite{baker2016designing, zoph2016neural}. Because each operation is associated with different layer-specific parameters which were implemented by hard-coded techniques, thus, this network representation strongly depends on expert knowledge.
    
\begin{figure}[!t]
    \begin{center}
      \includegraphics[width=\linewidth]{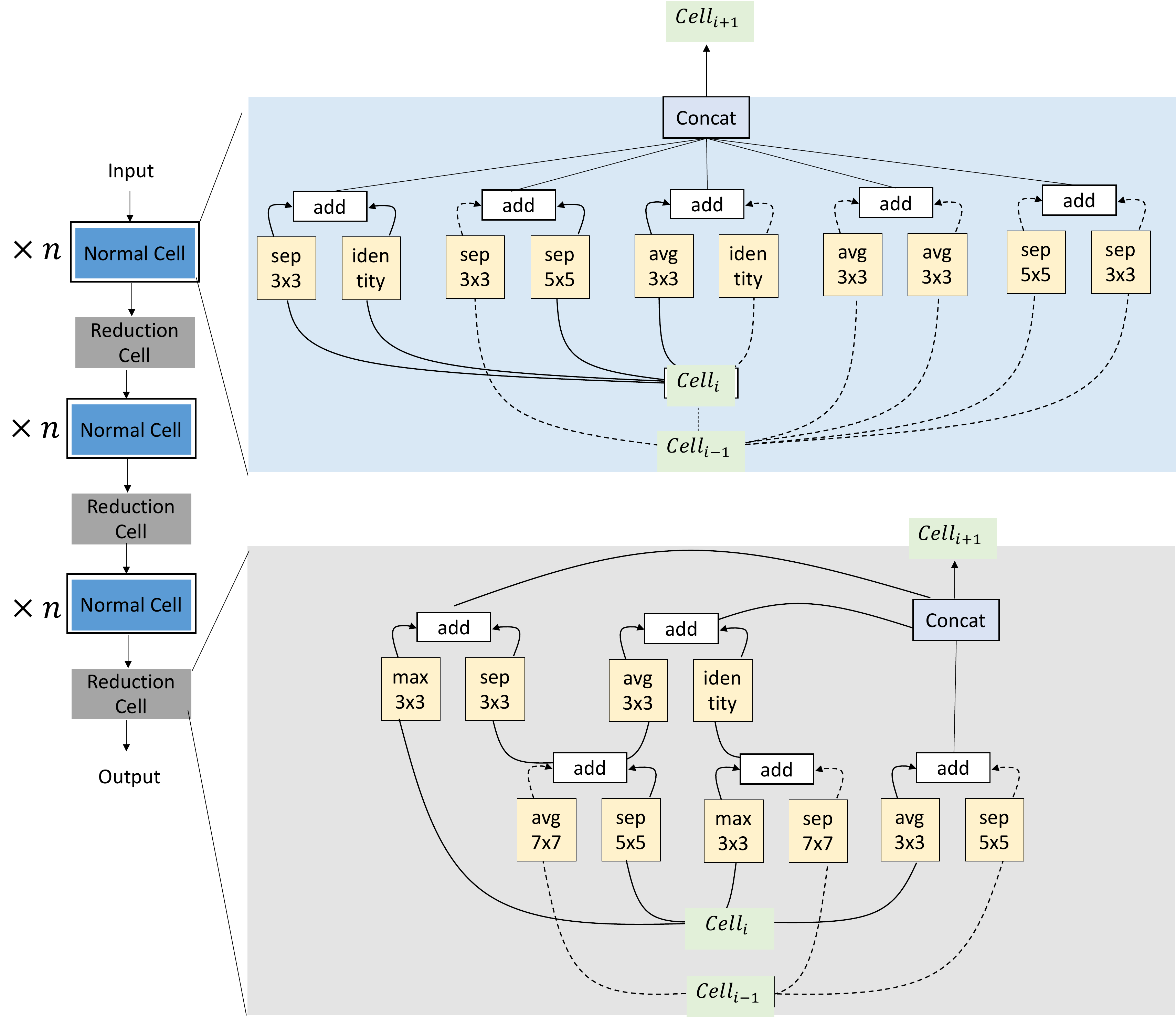}
    \end{center}
      \caption{Left: The overall structure of the \textit{search space} of NASNet with two cells: normal cell and reduction cell. The normal cell is repeated $\times n$ and then connected to a reduction cell. Right: Architecture of the best normal cell and reduction cells.}
    \label{fig:nasnet}
    \end{figure}

    To deal with too many nodes and edges in an entire architecture as well as to reduce the complexity of NAS search tasks, search spaces are usually defined over some smaller building block and learned meta-architecture to form a larger architecture \cite{elsken2019neural, wang2020m, elsken2020meta}. NASNet \cite{zoph2018learning} is one of the first works to make use of cell-based neural architecture. There are two types of cells in NASNet: normal cells and reduction cells. The former cells aim to extract advanced features while keeping the spatial resolution unchanged whereas the later cells aim to reduce the spatial resolution. A complete network contains many blocks, each block consists of multiple repeated normal cells followed by a reduction cell. An illustration of NASnet is shown in Fig.\ref{fig:nasnet} Leveraging NASNet, \cite{liu2018darts, real2019regularized, piergiovanni2019evolving, zhong2018practical} proposed similar cell-based search space but the reduce cells are eliminated as in Block-QNN \cite{zhong2018practical} and DPPNet \cite{ piergiovanni2019evolving} and replaced by pooling layers. In order to deal with U-Net-like encoder-decoder architectures, AutoDispNet \cite{saikia2019autodispnet} contains three types of cells: normal, reduction, and upsampling. At the encoder path, it comprises alternate connections of normal cells and reduction cells while the decoder path consists of a stack of multiple upsampling cells. More details of the cells from popular cell-based search spaces are theoretically and experimentally studied in \cite{shu2019understanding}. Instead of stacking one or more identical cells, FPNAS \cite{cui2019fast} considers stacking more diversity of blocks which aim to the improvement of neural architecture performance. Instead of a graph of operations, \cite{brock2017smash} consider a neural network as a system with multiple memory blocks which can read and write. Each layer operation is designed to: firstly, read from a subset of memory blocks; secondly, computes results; finally, write the results into another subset of blocks.

    Since the size of the search space is exponentially large or even unbounded, incorporating prior knowledge about properties well-suited for the task can reduce the size of the search space and simplify the search task. This could introduce a human bias, which may prevent the discovery of the optimal architectural building blocks that go beyond the current state-of-the-art.
    
\subsection{Search Strategy}
Given a search space, there are various search methods to select a configuration of network architecture for evaluation. The search strategy defines the way to explore the search space for the sake of finding high-performance architecture candidates. The most basic search strategies are grid search (i.e. systematically screening the search space) and random search (i.e randomly selecting architectures from the search space to be tested)\cite{bergstra2012random}. They are quite effective in practice for a small search space but they may fail with a large search space. Among hyperparameter optimization search methods, gradient-based approaches \cite{kandasamy2018neural} and Bayesian optimisation \cite{shahriari2015taking} based on Gaussian processes or Gaussian distribution has already proven its usefulness, specially in continuous space \cite{kandasamy2016gaussian, klein2017fast}. However, they may not work well with the discrete and high dimensionality space. To apply the gradient, gradient-based approaches transform the discrete search problem into a continuous optimization problem. With distribution assumption, Bayesian-based approaches rely on the choice of kernels. By contrast, evolutionary strategies \cite{real2019regularized, song2021enas} is more flexible and can be applied to any search space. However, evolutionary methods require to define a set of possible mutations to apply to different architectures. As an alternation, reinforcement learning is used to train a recurrent neural network controller to generate good architectures \cite{zoph2016neural, pham2018efficient}.
    
    The conventional NAS algorithms, which leverage either gradient-based or Bayesian optimization or evolutionary search or reinforcement learning, can be prohibitively expensive as thousands of models are required to be trained in a single experiment.
    
\begin{figure}[!t]
    \begin{center}
      \includegraphics[width=\linewidth]{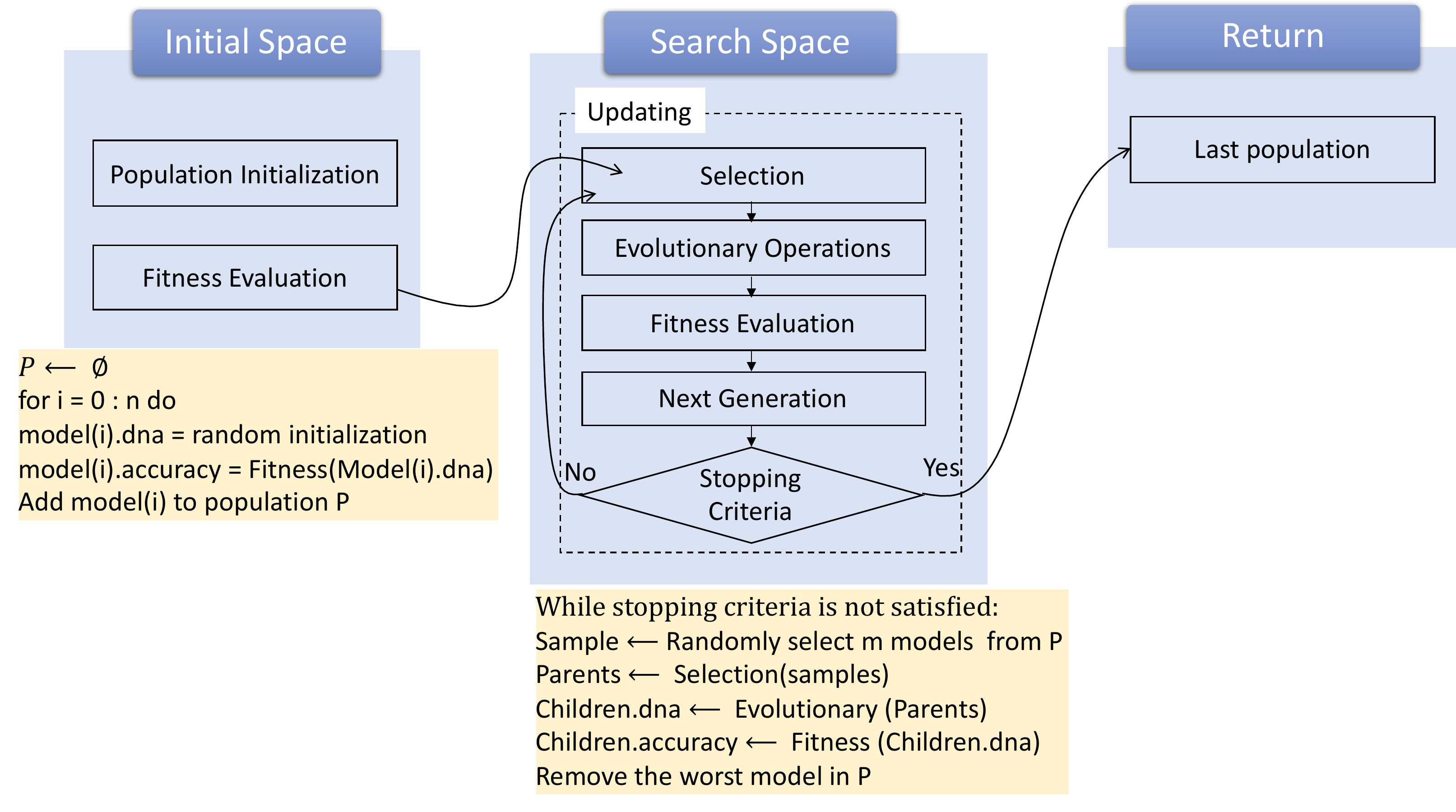}
    \end{center}
      \caption{Top: Flowchart of evolutionary strategy in NAS. Bottom: Pseudocode of evolutionary strategy in NAS.}
    \label{fig:enas}
    \end{figure}

    Compared to RL-based algorithms, gradient-based
    algorithms are more efficient although even gradient-based
    algorithms require the construction of supernet in advance,
    which also highly requires expertise. Due to the improper relationship for adapting to gradient-based optimization, both RL-based and gradient-based algorithms are often ill-conditioned architectures. Among all search strategies, evolutionary strategy (ES) NAS is the most popular, which can use some common approaches such as genetic algorithms \cite{sivanandam2008genetic}, genetic programming \cite{langdon2013foundations}, particle swarm optimization \cite{kennedy1995particle}. Fig.\ref{fig:enas} shows an illustration of the flowchart of an ES algorithm which takes place in the initial space and the search space sequentially. ES starts with a population being initialized within the initial space. In the population, each individual represents a solution (i.e., a DNN architecture) for NAS and it is evaluated by fitness process. After the initial population is fitness evaluated, the whole population starts the evolutionary process within the search space. In the evolutionary process, the population is updated by the selection and the evolutionary operators in each iteration, until the stopping criterion is met. Finally, a population that has finished the evolution is obtained. 
       
\subsection{Evaluation Strategy} For each hyperparameter configuration (from search strategy), we need to evaluate its performance. The evaluation can be benchmarked by either fully or partially training a model with the given hyperparameters, and subsequently measuring its quality on a validation set. Full training evaluation is a default method and it is known as the first generation of NAS evaluation strategy, which requires thousands of GPU days to achieve the desired result. To speed up the evaluation process, partial training methods make use of early-stopping. Hypernetworks \cite{brock2017smash, zhang2018graph}, network morphisms, \cite{cai2018path, jin2018efficient} and weight-sharing \cite{bender2018understanding, liu2018darts} are common NAS evaluation methods. Among these three methods, weight-sharing is least complex because it does not require training an auxiliary network while network morphisms are the most expensive as it requires on the order of 100 GPU days.

\begin{figure*}
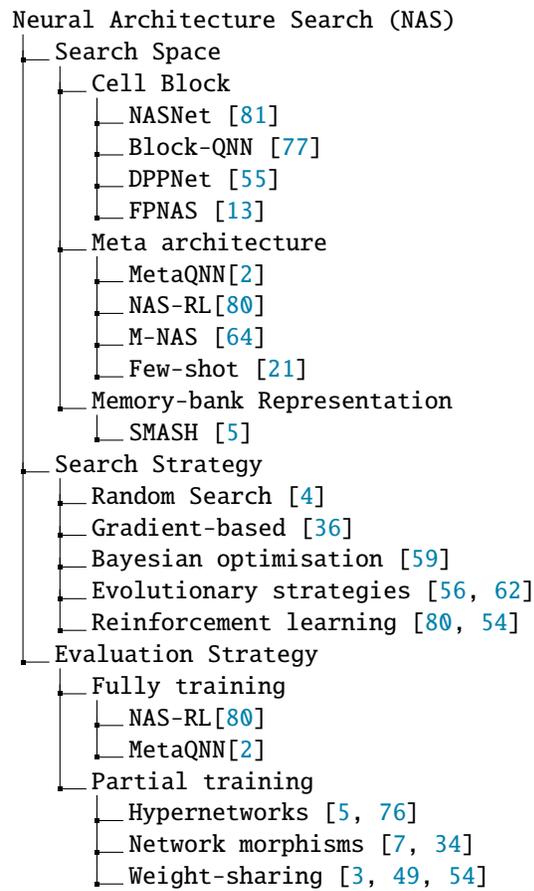

\dirtree{%
.1 Neural Architecture Search (NAS).
.2 Search Space.
.3 Cell Block.
.4 NASNet \cite{zoph2018learning}.
.4 Block-QNN \cite{zhong2018practical}.
.4 DPPNet \cite{ piergiovanni2019evolving}.
.4 FPNAS \cite{cui2019fast}.
.3 Meta architecture.
.4 MetaQNN\cite{baker2016designing}.
.4 NAS-RL\cite{zoph2016neural}.
.4 M-NAS \cite{wang2020m}.
.4 Few-shot \cite{elsken2020meta}.
.3 Memory-bank Representation.
.4 SMASH \cite{brock2017smash}.
.2 Search Strategy.
.3 Random Search \cite{bergstra2012random}.
.3 Gradient-based \cite{kandasamy2018neural}.
.3 Bayesian optimisation \cite{shahriari2015taking}.
.3 Evolutionary strategies \cite{real2019regularized, song2021enas}.
.3 Reinforcement learning \cite{zoph2016neural, pham2018efficient}.
.2 Evaluation Strategy.
.3 Fully training.
.4 NAS-RL\cite{zoph2016neural}.
.4 MetaQNN\cite{baker2016designing}.
.3 Partial training.
.4 Hypernetworks \cite{brock2017smash, zhang2018graph}.
.4 Network morphisms \cite{cai2018path, jin2018efficient}.
.4 Weight-sharing \cite{bender2018understanding, liu2018darts, pham2018efficient}.
}
\caption{Summary NAS architecture regarding search space, search strategy, and evaluation strategy.}
\label{fig:sum_NAS}
\end{figure*}

NAS approaches categorized by search space, search strategy, and evaluation strategy are summarized in Fig.\ref{fig:sum_NAS}.

\section{NAS for Medical Imaging}
The recent breakthroughs in NAS have motivated various applications in medical images such as segmentation, classification, reconstruction, etc. Starting from NASNet \cite{zoph2018learning}, many novel search spaces, search strategies, and evaluation strategies have been proposed for biomedical images. The following sections will detail recent efficient NAS for medical imaging applications.

\subsection{NAS for Medical Image Classification}
In image classification, the  search space can be divided into either network topology level \cite{xie2019exploring, fang2020densely}, which perform search on the network topology; or cell level \cite{liu2018progressive, liu2018darts, pham2018efficient, real2019regularized}, which focus on searching optimal cells and apply a predefined network topology. NASNet \cite{zoph2018learning} is considered as one of the first successful NAS architectures for image classification. The overall architecture of NASNet together with its normal cells and reduction cells are shown in Fig.\ref{fig:NAS}. In NASNet, the normal cell returns a feature map of the same dimension whereas the reduction cell returns a feature map where the feature map height and width are reduced by a factor of two. Both normal cell and reduction cell are searched by a controller, which is fashioned from Recurrent Neural Network (RNN). 

\begin{figure}[!t]
\begin{center}
  \includegraphics[width=\linewidth]{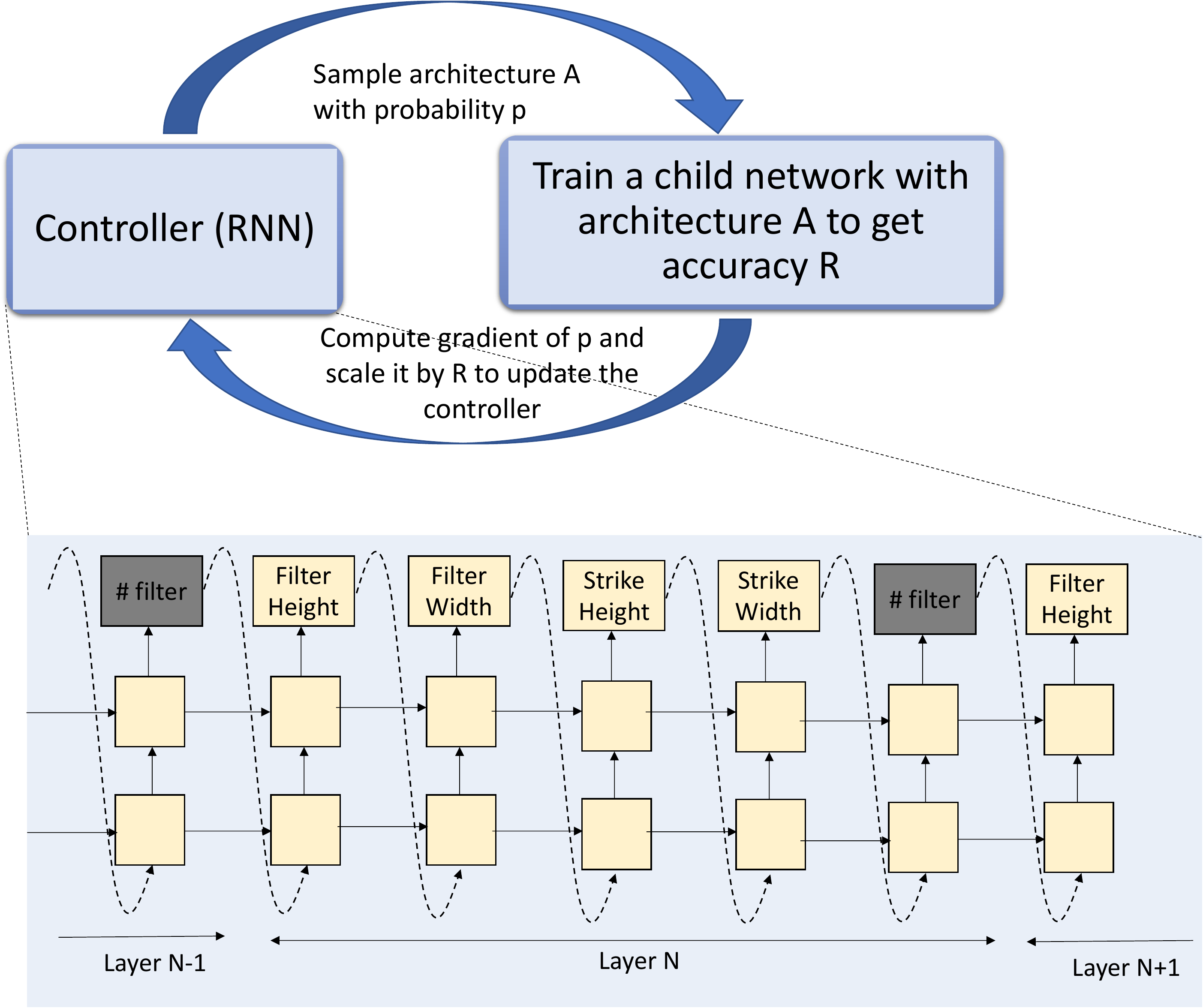}
\end{center}
  \caption{Top: An overview of Neural Architecture Search. Botton: NAS Controller by RNN}
\label{fig:NAS_Le}
\end{figure}

In this section, we take the frontier NAS approach \cite{zoph2016neural} developed by Google Brain as an instance to show how to employ NAS into image classification. The network architecture of NAS \cite{zoph2016neural} is shown in Fig.\ref{fig:NAS_Le}(top) where the controller is defined as RNN based on Long Short Term Memory (LSTM) \cite{hochreiter1997long} is shown in Fig.\ref{fig:NAS_Le}(bottom). The RNN controller is responsible for generating new architectural hyper-parameters of CNNs, and is trained using REINFORCE \cite{williams1992simple}. In NAS, controller predicts the parameters $\theta_C$ corresponding to filter height, filter width, stride height, stride width, and number of filters for one layer and repeats. Every prediction is carried out by a softmax classifier and then fed into the next time step as input. The parameters $\theta_C$ are optimized in order to maximize the expected validation accuracy of the proposed architectures. After the controller predicts a set of hyper-parameters $\theta_C$, a neural network with the specified configuration is built, i.e. a child model, and trained to convergence on a dataset (CIFAR-10 is used). In this architecture, the child network accuracy $R$ is utilised as a reward to train the controller under a reinforcement learning mechanism. In NAS, each gradient update to the controller parameters $\theta_C$ corresponds to training one child network to convergence. An upper threshold on the number of layers in the CNNs is used to stop the process of generating new architectures. Each network proposed by the RNN controller is trained on CIFAR10 dataset for fifty epochs with the use of vast computational resources (450 GPUs for 3-4 days for a single experiment). The search space contains 12,800 architectures. 

Based on NAS, many effective NAS approaches on image classification have been proposed such as evolution-based NAS \cite{real2019regularized} (i.e. using evolution algorithms to simultaneously optimize topology alongside with parameters), ENAS  \cite{pham2018efficient} (i.e. sharing of parameters among child models), DARTS \cite{liu2018darts} (i.e. formulating the task in a differentiable manner to shorten the search within four GPU days), GDAS \cite{dong2019searching} (i.e. enabling the search speed in four GPU hours), ProxylessNAS \cite{cai2018proxylessnas} (i.e. process on a large-scale
target tasks and the target hardware platforms)

Inspired by the successes of NAS in computer vision, i.e. image classification, there are several attempts in medical image classification that employed NAS techniques. Leveraged by \cite{elsken2017simple}, which uses the hill-climbing algorithm with the network morphism transformation to search for the architectures, Kwasigroch et al.,\cite{kwasigroch2020neural} proposed a malignant melanoma detection for skin lesion classification. In this method, the hill-climbing algorithm can be interpreted as a simple evolutionary algorithm with only a network morphism operation. 
Adopt AdaNet framework \cite{cortes2017adanet} as the NAS engine, \cite{dai2020optimize} proposed Adanet-NAS to optimize a CNN model for three classes fMRI signal classification.

\subsection{NAS for Medical Image Segmentation}
Medical image segmentation faces some unique challenges such as lacking annotated data, inhomogeneous intensity, and vast memory usage for processing 3D high-resolution images. 3DUnet \cite{ronneberger2015u}, VNet\cite{milletari2016v} are the first 3D networks designed for medical image segmentation. Later, many other effective CNN-based 3D networks for medical image segmentation such as cascade-Unet \cite{le2021multi}, UNet++ \cite{zhou2019unet}, Densenet \cite{huang2017densely}, H-DenseUNe \cite{li2018h}, NN-UNet \cite{isensee2019automated} have been proposed.

In computer vision, NAS has mainly solved image classification and a few recent works recently applied NAS to image segmentation such as FasterSeg\cite{chen2019fasterseg}, Auto-DeepLab \cite{liu2019auto}. 

\begin{figure}
   \includegraphics[width=\textwidth]{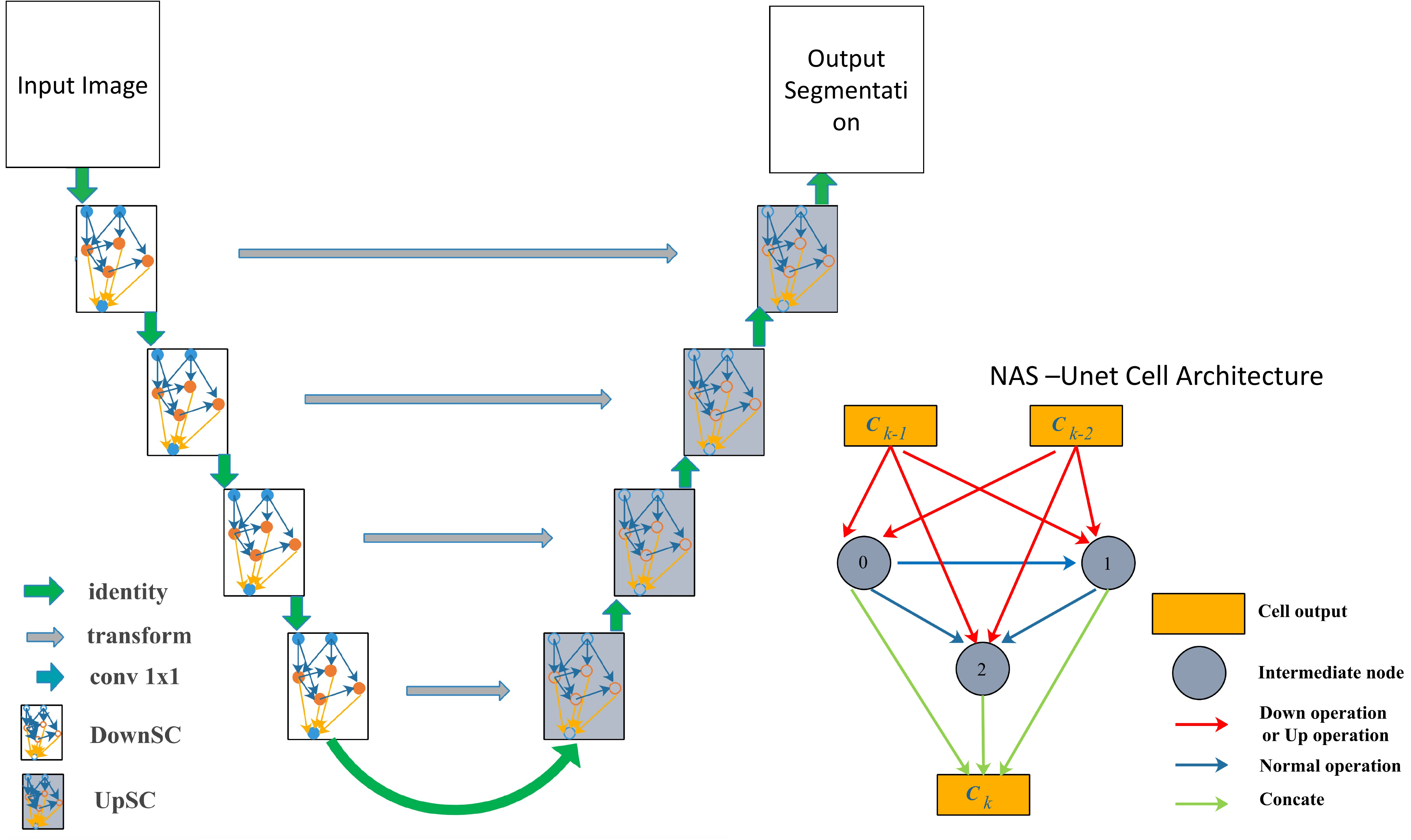}
  \caption{Left: The U-like backbone of NAS-Unet architecture, the rectangle represents cell architectures need to search. The green arrow merely represents the flow of the feature map (input image). The gray arrow is a transform operation that belongs to UpSC and is also automatically searched. Right: NAS-Unet cell architecture. The red arrow indicates a down operation, the blue arrow indicates the normal operation and the green arrow represents a concatenate operation. Courtesy of \cite{weng2019unet}. }
\label{fig:NAS-Unet}
\end{figure}

\begin{figure}
  \begin{minipage}[c]{0.5\textwidth}
   \includegraphics[width=\textwidth]{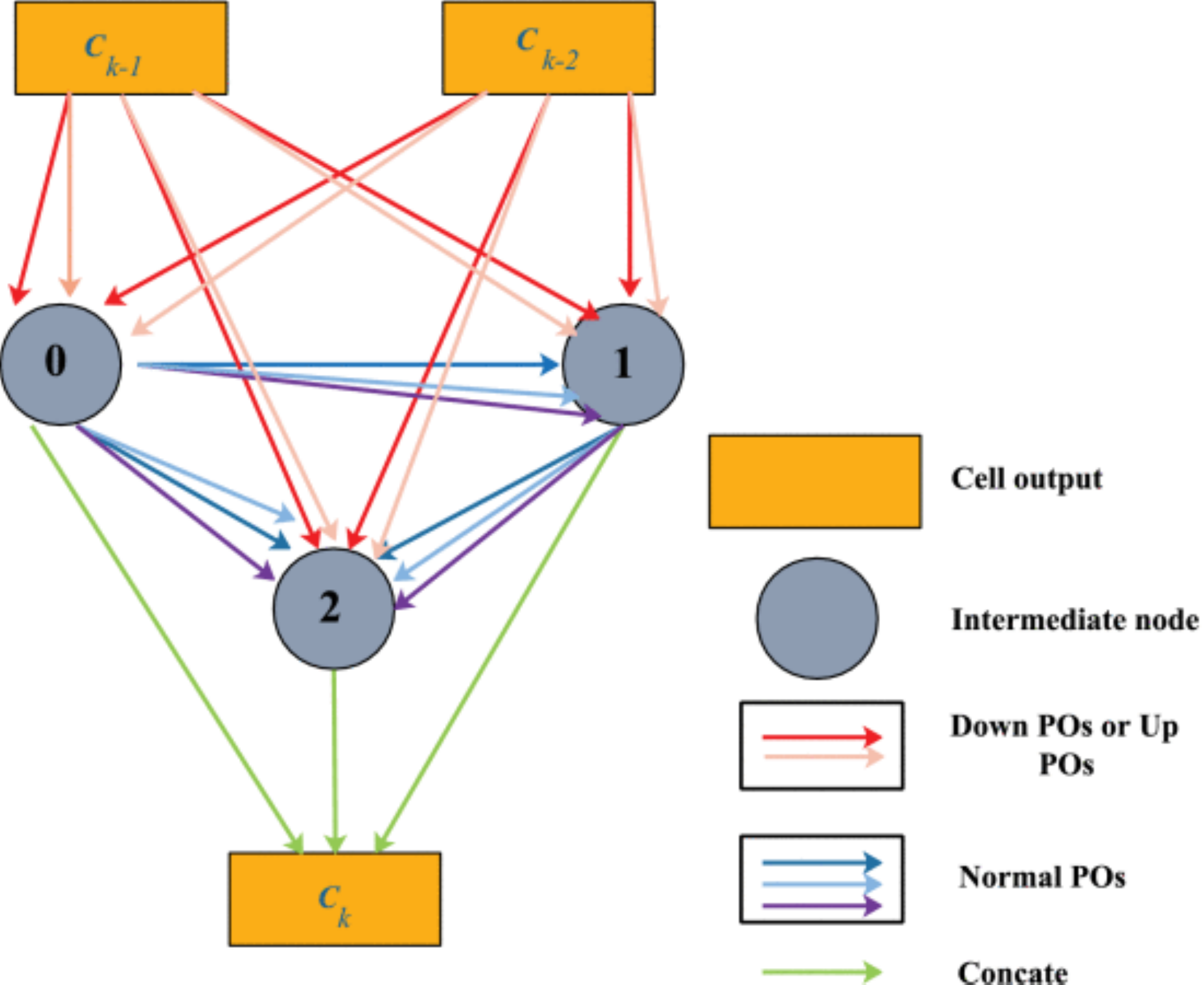}
  \end{minipage}\hfill
  \begin{minipage}[c]{1.0\textwidth}
    \includegraphics[width=\textwidth]{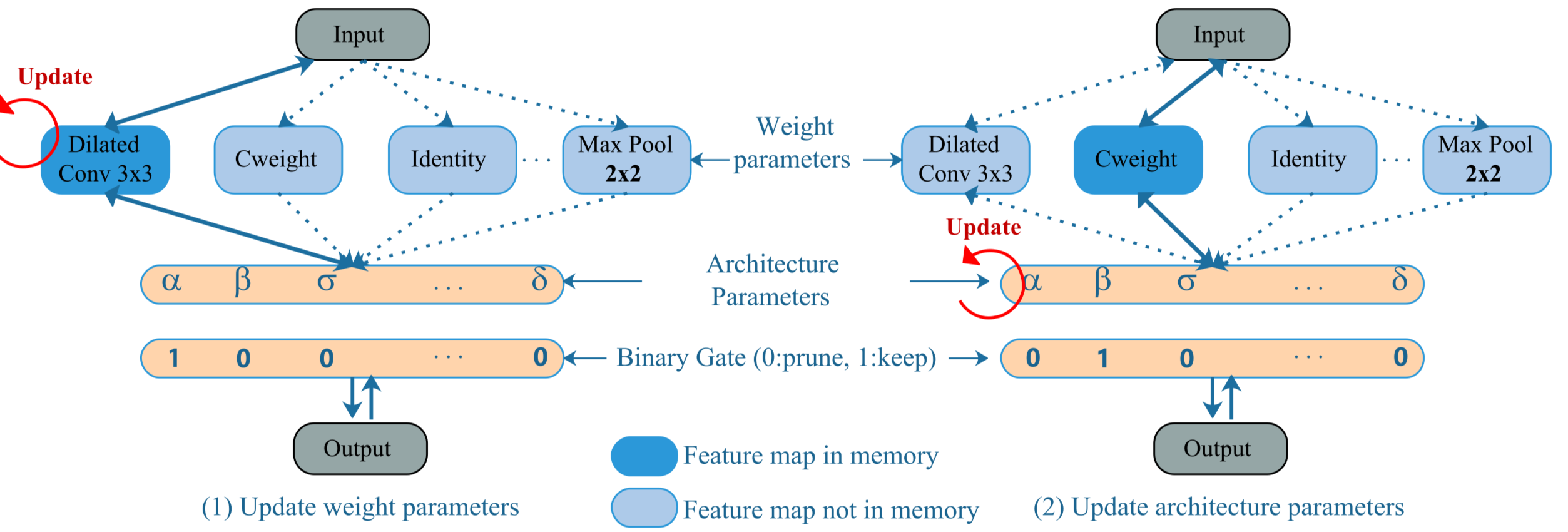}
  \end{minipage}
  \caption{Top: Over-parameterized cell architecture a NAS-Unet. Each edge is associated with N candidate operations from different primitive operation sets. Bottom: Update strategy by ProxylessNAS \cite{cai2018proxylessnas}. Courtesy of \cite{weng2019unet}.}
\label{fig:over_params_cell}
\end{figure}

In medical analysis, accurate segmentation of medical images is a crucial step in computer-aided diagnosis, surgical planning, and navigation which have been applied in a wide range of clinical applications. Great achievements have been made in medical segmentation thanks to the recent breakthroughs in deep learning, such as 3D-UNet and NN-UNet. However, it remains very difficult to address some challenges such as extremely small objects with respect to the whole volume, weak boundary, various location, shape, and appearance. Furthermore, volumetric image segmentation is extremely expensive to train, thus, it is difficult to attain an efficient 3D architecture search. NAS-Unet \cite{weng2019unet} and V-NAS \cite{zhu2019v} are two of the first NAS architectures for medical segmentation. NAS-Unet is based on U-like/V-net backbone (i.e. Densenet implementation \cite{jegou2017one} in NAS-NET and V-net\cite{milletari2016v} in V-NAS) with two types of cell architectures called DownSC and UpSC as given in Fig.\ref{fig:NAS-Unet} (left). NAS-UNet is based on cell-block building search space and the search cell contains three types of primitive operation: Down PO (average pooling, max pooling, down cweight, down dilation conv., down depth conv. and down conv.), Up PO (up cweight, up depth conv., up conv., up dialtion conv.) and Normal PO (identity, cweight, dilation conv., depth conv., conv.) as shown in Fig.\ref{fig:NAS-Unet} (right). In NAS-Unet, both DownSC and UpSC are simultaneously updated by a differential architecture strategy during the search stage. As given in Fig.\ref{fig:NAS-Unet}(a), the NAS-Unet contains $L_1$ cells in the encoder path o learn the different level of semantic context information and $L_1$ cells in the decoder path to restore the spatial information of each probability. Compared to Densenet \cite{huang2017densely}, NAS-Unet replaces the convolution layers with these cells and moves up-sampling operation and down-sample operation into the cells. Leveraged by DARTS \cite{liu2018darts}, the NAS-Unet constructs an over-parameterized network by Eq. \ref{eq:C}.
\begin{equation}
    C(e_1=MixO_1,\dots,e_E=MixO_E)
\label{eq:C}
\end{equation},
where each edge is a mixed operation that has $N$ parallel paths, denoted as MixO, as shown in Fig.\ref{fig:over_params_cell} (left). The output of a mixed operation $MixO$ is defined based on the output of its $N$ paths as in Eq.\ref{eq:Mi}
\begin{equation}
    MixO(x)=\sum _{(i=1)}^{N}w_{i}o_{i}(x)
\label{eq:Mi}
\end{equation}
To save memory during the update strategy, NAS-Unet makes use of ProxylessNAS \cite{cai2018proxylessnas} as shown in Fig.\ref{fig:over_params_cell}(right). By using ProxylessNAS, NAS-Unet update of one of the architecture parameters by gradient descent at each step. To obtain that objective, NAS-Unet first freezes the architecture parameters and stochastically sample binary gates for each batch of input data. NAS-Unet then updates parameters of active paths via standard gradients descent on the training dataset as shown in Fig.\ref{fig:over_params_cell} (right). These two update steps are performed in an alternative manner. 
\textit{}

\begin{figure}[!ht]
   \includegraphics[width=1.0\textwidth]{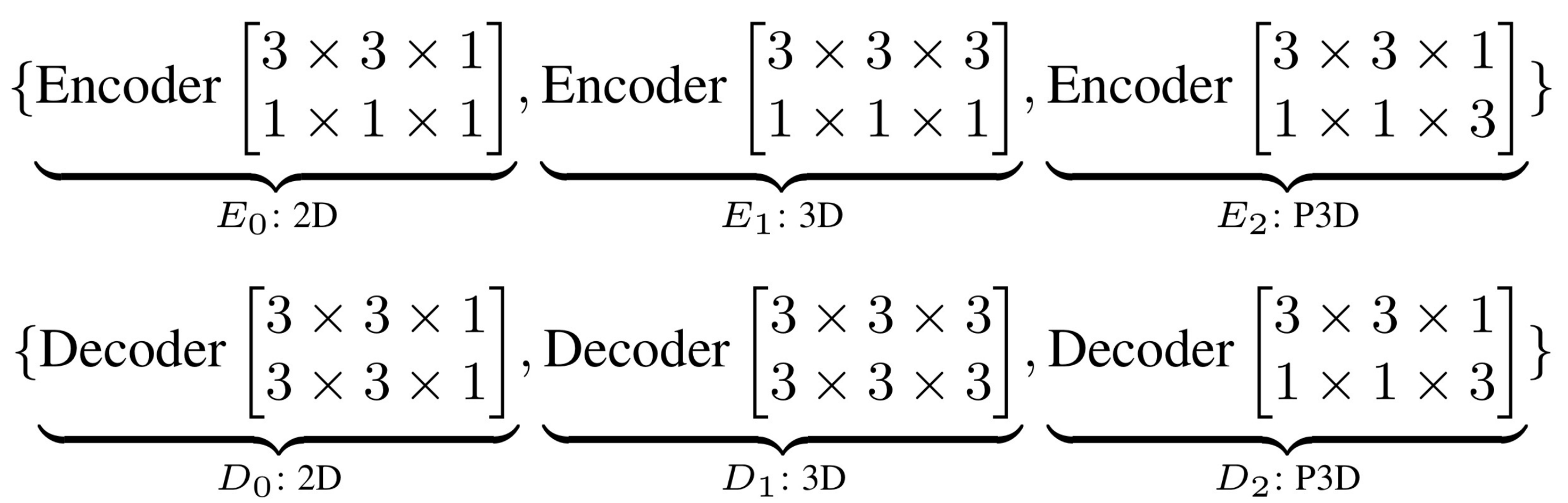}
   \caption{Encoder cell and decoder cell defined in V-NAS \cite{zhu2019v}.}
   \label{fig:encoder_decoder_vnas}
\end{figure}

In addition to NAS-Unet, V-NAS \cite{zhu2019v} is another NAS cell-blocked-based NAS architecture and based on V-Net\cite{milletari2016v} network design. In V-NAS, a cell is defined as a fully convolutional module composing of several convolutional (Conv+BN+ReLU) layers, which is then repeated multiple times to construct the entire neural network. Corresponding encoder path and decoder path, V-NAS consists of encoder cell and decoder cell which are chosen between 2D, 3D, or Pseudo-3D (P3D) as shown in Fig.\ref{fig:encoder_decoder_vnas}. V-NAS is designed with ResNet-50 in the encoder path and pyramid volumetric pooling (PVP) in the decoder path.

\cite{zhu2019v} designed a search space consisting of both 2D, 3D, and pseudo-3D (P3D) operations, and let the network itself select between these operations at each layer.

\begin{figure*}
   \includegraphics[width=\textwidth]{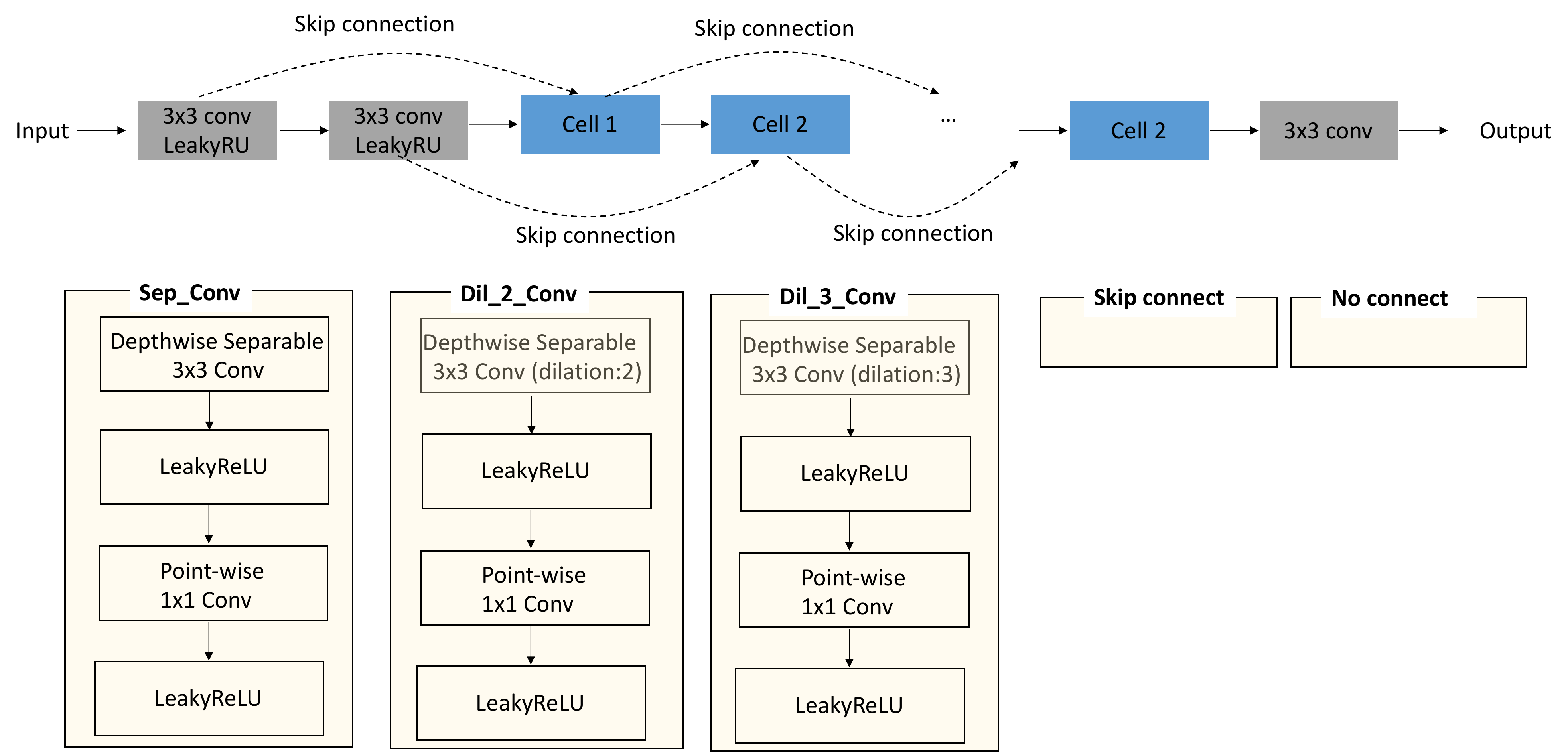}
  \caption{Top: The reconstruction module used in NAS-based reconstruction \cite{yan2020neural}. Down: Candidate layer operations in search space.}
\label{fig:reconstruction_cell}
\end{figure*}

Different from NAS-Unet \cite{weng2019unet} and V-NAS \cite{zhu2019v} which search cells and apply them to a U-Net/V-Net like architecture, C2FNAS \cite{yu2020c2fnas} searches 3D network topology in a U-shaped space and then searches the operation for each cell. The search procedure contains two stages corresponding to macro level (i.e. defining how every cell is connected to each other) and micro-level ( assigning an operation to each node). Thus, a network is constructed from scratch in a macro-to-micro manner under two stages which aim to relieve the memory pressure and resolve the inconsistency problem between the search stage and
deployment stage. MS-NAS \cite{yan2020ms} applied PC-DARTS \cite{xu2019pc} and Auto-DeepLab’s formulation to 2D medical images MS-NAS is defined with three types of cells: expanding cells (i.e. expands and up-samples the scale of feature map), contracting cells (i.e. contracts and down-samples the scale of feature map;), and non-scaling cells (i.e. keeps the scale of feature map constant) to automatically determine the network backbone, cell type, operation parameters, and fusion scales. Recently, BiX-NAS\cite{wang2021bix} searches for the optimal bi-directional architecture by recurrently skipping multi-scale features while discarding insignificant ones at the same time. 

One of the current limitations of NAS is that it lacks corresponding baseline and sharable experimental protocol. Thus, it is difficult to compare between NAS search algorithms. In this section, we make a comparison based on the performance on a particular dataset without knowing where is the performance gains come from.

\subsection{NAS for Other Medical Image Applications}

Inspired by continuous relaxation
of the architecture representation of DARTS \cite{liu2018darts} with differentiable search, \cite{yan2020neural} proposed NAS-based reconstruction which searches for the internal structure of the cells. The reconstruction module is a stack of cells as shown in Fig.\ref{fig:reconstruction_cell} (left) where the first and the last common $3 \times 3$ convolutional layer. For each cell, it maps the output tensors of previous two cells to construct its output by concatenating two previous cells with a parameter representing the relaxation of discrete inner cell architectures. The search space contains three operators defined as in the Fig.\ref{fig:reconstruction_cell} (right). The inner structure of cells is search through DARTS \cite{liu2018darts}. 

A similar NAS-based MRI reconstruction network was introduced by EMR-NAS \cite{huang2020enhanced} where the search space contains eight different cells with the same kernel size $3 \times 3$  but different dilation rate and the connection between them

Besides classification, segmentation, and reconstruction, lesion detection is another important task in medical analysis. In addition to TruncatedRPN balances positive and negative data for false-positive reduction; ElixirNet \cite{jiang2020elixirnet} proposed Auto-lesion Block (ALB) to locate the tiny-size lesion by dilated convolution with flexible receptive fields. The search space for ALB contains nine operators i.e. $3\times 1$ and $1\times 3$ depthwise-separable conv, $3\times 3$ depthwise-separable conv, $5\times 5$ depthwise-separable conv, $3\times 3$ atrous conv with dilate rate 3, $5\times 5$ atrous conv with dilate rate 5, average pooling, skip connection, no connection, non-local. All operations are of stride 1 and the convolved feature
maps are padded to preserve their spatial resolution. Among all operators, non-local operator aims to encode semantic relation between region proposals that is relevant to lesion detection. The cell is searched by DARTS \cite{liu2018darts}.

NAS is also employed to localize multiple uterine  standard plan (SP) in 3D Ultrasound (US) simultaneously by Multi-Agent RL (MARL) framework \cite{yang2021searching}. In MARL, the optimal agent for each plane is obtained by one-shot NAS \cite{guo2020single} to avoid time-consuming. The search strategy is based on GDAS \cite{dong2019searching} which search by gradient descent and only updates the sub-graph
sampled from the supernet in each iteration. In MARL, the search space contains eight cells (five normal cells and three reduce cells) and each agent has its own four cells (three normal cells and one reduce cells). The cell search space consists of ten operations including none, $3\times 3$ conv., $5\times 5$ conv., $3\times 3$ dilated conv.,
$5\times 5$ dilated conv., $3\times 3$ separable conv., $5\times 5$ separable conv.,
$3\times 3$ max pooling, $3\times 3$ avg pooling, and skip-connection.

\section{Meta-Learning in NAS}

NAS has made remarkable progress in many tasks in both medical imaging and computer vision; however, NAS is not only computation resource consumption but also requires a large amount of annotated data which is one of the biggest challenges in medical imaging. Furthermore, most of the existing NAS methods search for a well-performing architecture for a single task. Meanwhile prior meta-learning approaches \cite{finn2017model, nichol2018first, antoniou2018train, finn2018probabilistic} compute meta-learning weights to learn new tasks from just a few examples, but their model is trained with a fixed neural architecture. 

\begin{figure*}
 \includegraphics[width=\textwidth]{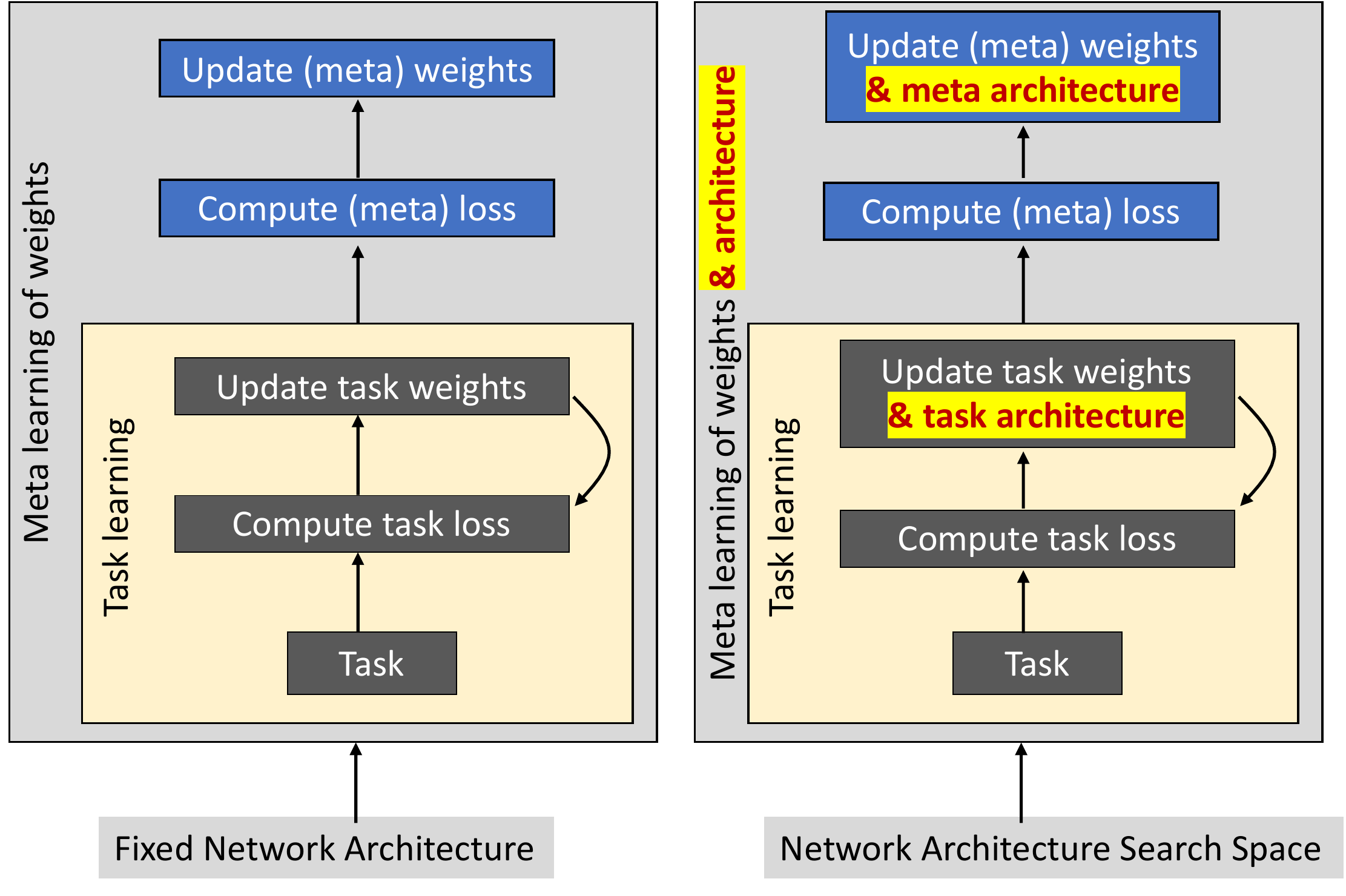}
  \caption{Comparison between meta-learning with a fixed architecture (left) and meta-learning with NAS (right). The main difference is updating process (highlighted). In meta-learning, only network weights are updated whereas both network architecture and its corresponding weights are updated in meta-learning with NAS. }
\label{fig:compare-meta}
\end{figure*}

Based on the observation that NAS can be seen as an instance of hyper-parameter meta-learning, there have been some works that combine NAS and meta-learning to address both new tasks and flexible network architecture. The comparison between meta-learning only and meta-learning with NAS is illustrated in Fig. \ref{fig:compare-meta}. Fig. \ref{fig:compare-meta} (left) is meta-learning algorithm that takes a fixed network architecture as its input and it updates network weights. Whereas Fig. \ref{fig:compare-meta} (right) is meta-learning  with NAS which takes a network architecture search space as its input and simultaneously updates both network architecture with its corresponding weights. 

To train a network on multiple tasks, Wong et al. \cite{wong2018transfer} train an AutoML system \cite{hutter2019automated} via RL with utilization of transfer learning. In Wong et al. \cite{wong2018transfer}, hyper-parameters are searched for a new task while the architecture is chosen as one of the pre-defined ones. Zoph et al. \cite{zoph2018learning} make use of NAS \cite{zoph2016neural} and propose a design of a new search space, called NASNet search space, to search for a good architectures on a small dataset and then transfer across a range of data and computational scales. Kim et al. \cite{kim2018auto} wrap NAS around meta-learning by adopting progressive NAS \cite{liu2018progressive} to few-shot learning. In Kim et al. \cite{kim2018auto}, the model is trained from scratch in every iteration of the NAS algorithm, thus, it is computation consumption. In a concurrent work \cite{lian2019towards}, Lian et al. propose a meta-learning-based transferable neural architecture search method to generate a meta-architecture, which can adapt to new tasks through a few gradient descent steps. In Lian et al. \cite{lian2019towards}, the model is trained in every task-dependent architecture, thus, it is time consuming. By incorporating gradient-based NAS and gradient-based meta-learning, Elsken et al. \cite{elsken2020meta} propose METANAS which can adapt an architecture to new tasks using a small number of samples with only several gradient descent steps. Elsken et al. \cite{elsken2020meta} not only meta-learns network weights for a given fixed architecture, but also meta-learns architecture as shown in  Fig.\ref{fig:compare-meta}(right). That means, both network weights and network architecture are meta-learned simultaneously.

To optimize both network architecture and its corresponding weights at once and adapt them to a new task, METANAS utilizes both task-learner (e.g DARTS~\cite{liu2018darts}) and meta-learner (e.g. REPTILE \cite{nichol2018first}). Let denote $\alpha$ and $\omega$ as meta-learned architecture and its corresponding meta-learned weights. METANAS objective aims to quickly adapt $\alpha$ and $\omega$ to a new task $\mathcal{T}_i$ with a few samples $\mathcal{T}_i = {D^{\mathcal{T}_i}_{train}, D^{\mathcal{T}_i}_{val}}$, where training data $D^{\mathcal{T}_i}_{train}$ is sampled from training task distribution $p_{train}$. The optimal architecture and its corresponding optimal weights of a task $\mathcal{T}_i$ are denoted as $\alpha^*(\mathcal{T}_i)$ and $\omega^*(\mathcal{T}_i)$. To archive such goals, the METANAS loss is defined as follows:
\begin{equation}
\begin{split}
\mathcal{L}(\alpha, \omega, p_{train}, \phi_k) & =  \sum_{\mathcal{T}_i}{\mathcal{L}_{\mathcal{T}_i}(\phi_k(\alpha, \omega, D^{\mathcal{T}_i}_{train}), D^{\mathcal{T}_i}_{val})} \\
& = \sum_{\mathcal{T}_i}{\mathcal{L}_{\mathcal{T}_i}}{((\alpha^*(\mathcal{T}_i), \omega^*(\mathcal{T}_i)), D^{\mathcal{T}_i}_{val})}
\end{split}
\end{equation}

, where $\phi_k$ is task-learner at $k^{th}$ updating iteration. NAS by DARTS \cite{liu2018darts} is used as task-learner and applied to update both architecture and its corresponding weight, i.e., $[\alpha; \omega]$. $\alpha$ and $\omega$ are updated with their own learning rate $\lambda^{task}_\alpha$ and $\lambda^{task}_\omega$ to obtain the optimal architecture $\alpha^*(\mathcal{T}_i)$ and its corresponding optimal weights $\omega^*(\mathcal{T}_i)$ . The loss $\mathcal{L}$ is differentiable with respect to both $\alpha$ and $\omega$. The loss $\mathcal{L}_{\mathcal{T}_i}$ is the loss of particular task $\mathcal{T}_i$ and based on task loss.

\begin{algorithm}[t]
\caption{METANAS learning procedure with task-learner DARTS \cite{liu2018darts} and meta-learner REPTILE \cite{nichol2018first}.}
\label{algo:metanas}
\hrule
\begin{algorithmic}[1]
\algrenewcommand\algorithmicrequire{\textbf{Input: }}
\algrenewcommand\algorithmicensure{\textbf{Output: }}

\Require \\ Distribution over taks $p(\mathcal{T})$ \\ task-learner at k iteration $\phi^k$. \\
meta-learner $\theta_\alpha$, $\theta_\omega$
\Ensure meta-learned architecture $\alpha$, meta-learned weights $\omega$.

\State \textbf{Initialize}: meta-learned architecture $\alpha$, meta-learned weights $\omega$.

\State \While {not converge}
    \State Sample tasks $\mathcal{T}_i$ from $p(\mathcal{T})$
    \For{all $\mathcal{T}_i$ }
        \State $\alpha^{\mathcal{T}_i} \gets \alpha$
        \State $\omega^{\mathcal{T}_i} \gets \omega$
        \For {j = 1, ..., k }
        \Comment{{updating iteration}}
            \State $\alpha^{\mathcal{T}_i} \gets \phi^k(\alpha, \omega, D^{\mathcal{T}_i}_{train}, \lambda^{task}_\alpha)$ \Comment{{update architecture by a task-learner $\phi^k$ with learning rate $\lambda^{task}_\alpha$}}
            \State $\omega^{\mathcal{T}_i} \gets \phi^k(\alpha, \omega, D^{\mathcal{T}_i}_{train}, \lambda^{task}_\omega)$ \Comment{{update architecture by a task-learner $\phi^k$ with learning rate $\lambda^{task}_\omega$}}
        \EndFor
    \EndFor
    \State $\alpha \gets \alpha + \theta_\alpha(\alpha^{*\mathcal{T}_i}, \omega^{*\mathcal{T}_i}, \lambda^{meta}_{\alpha})$ \Comment{{update architecture by a meta-learner $\theta_\alpha$ with learning rate $\lambda^{meta}_\alpha$}}
    \State $\omega \gets \omega + \theta_\omega(\alpha^{*\mathcal{T}_i}, \omega^{*\mathcal{T}_i}, \lambda^{meta}_{\omega})$ \Comment{{update weights by a meta-learner $\theta_\omega$ with learning rate $\lambda^{meta}_\omega$}}
\EndWhile
\end{algorithmic}
\end{algorithm}

After finding the optimal architecture $\alpha^*(\mathcal{T}_i)$ and its optimal weights $\omega^*(\mathcal{T}_i)$ for a particular task $\mathcal{T}_i$, the meta-learned architecture $\alpha$ and the meta-learned weights $\omega$ are updated by a meta-learned $\theta_\alpha$ and $\theta_\omega$ with learning rates of $\lambda^{meta}_\alpha$ and $\lambda^{meta}_\omega$. REPTILE \cite{nichol2018first} is used as meta-leaner in METANAS. The entire METANAS learning process is described in Algorithm \ref{algo:metanas}.

While most of the existing meta-learning-based NAS methods focus more on a few shot learning with a small number of labeled data, there is no annotated data available in some cases. Liu et al. \cite{liu2020labels} propose an unsupervised neural architecture search (UnNAS) to explore whether labels are necessary for NAS. It is resulting that NAS without labels is competitive with those with labels; therefore, labels are not necessary for NAS. 

\section{Future Perspectives}
Artificially neuron networks (ANN) have made breakthroughs in many medical fields, including  recognition, segmentation, detection, reconstruction, etc. Compared with ANN, NAS is still in the initial research stage even NAS has become a popular subject in the area of machine-learning science. Commercial services such as Google’s AutoML and open-source libraries such as Auto-Keras make NAS accessible to the broader machine learning environment. At the current stage of development, NAS-based approaches focus on improving image classification accuracy, reducing time consumption during the search for a neural architecture. There are some challenges and future perspectives discussed as follows:

\textbf{Search space}: There are various effective search spaces; however, they are based on human knowledge and experience, which inevitably introduce human bias. Balancing between freedom of neural architecture design, search cost and network performance in NAS-based approach is an important future research direction. For example, to reduce the search space as much as possible while also improving network performance, NASNet \cite{zoph2018learning} proposes a modular search space that was later widely adopted. However, this comes at the expense of the freedom of neural architecture design. Thus, general, flexible, and human bias-free search space are critical requirements. To minimize human bias, AutoML-Zero \cite{elsken2020meta} applies the evolution strategy (ES) and designs two-layer neural networks based on basic mathematical operations (cos, sin, mean, st). 

\textbf{Robustness}: Even though NAS has been proven effective in many datasets, it is still limited when dealing with a dataset that contains noise, adversarial attacks or open-set dataset. Some efforts have been proposed to boost the NAS's robustness such as Chen et al. \cite{chen2020robustness} proposed a loss function for noise tolerance. Guoet al. \cite{guo2020meets} explored the intrinsic impact of network architectures on adversarial attacks.

\textbf{Learn new data}: Most of the existing NAS methods can search an appropriate architecture for a single task. To search for a new architecture on a new task, a suggested solution is to combine meta-learning into NAS \cite{pasunuru2019continual, elsken2020meta, lian2019towards}. For example \cite{lian2019towards} proposed a transferable neural architecture search to generate a meta-architecture, which can adapt to new tasks and new data easily while \cite{elsken2020meta} applying NAS into few-shot learning. On learning new data, recent work by \cite{liu2020labels} proposed unsupervised neural architecture search (UnNAS) shows that an architecture searched without labels is competitive with those searched with labels.

\textbf{Reproducibility}: Most of the existing NAS methods have many parameters that need to be set manually at the implementation level which may be described in the original paper. In addition to the design of the neural architecture, configure non-architecture hyperparameters (e.g. initial learning rate, weight decay, drop out ratio, optimizer type, etc) also time consumption and strongly affects network performance. Jointly search of hyperparameters and architectures has been taken into consideration; however, it just focuses on small data sets and small search spaces. Recent research AutoHAS \cite{dong2020autohas}, FBNet \cite{dai2020fbnetv3} show that jointly search of hyperparameters and architectures has great potential. Furthermore, reproduce NAS requires a vast of resource. With the rise of NAS-based techniques, it is now possible to produce state-of-the-art ANNs for many applications with relatively low search time consumption (a few 10 GPU-days rather than 1000 GPU-days). The future NAS-based approach should be able to address various problems across domains i.e. tasks: segmentation, object detection, depth estimation, machine translation, speech recognition, etc; computing platform: server, mobile, IOT and CPU, GPU, TPU; sensor: camera, lidar, radar, microphone; objectives: accuracy, parameter size, MACs, latency, energy. Instead of wasting much time on the model evaluation, some datasets (e.g NAS-Bench-101  \cite{ying2019bench}, NAS-Bench-201 \cite{dong2020bench}) support NAS researchers to focus on the design of optimization algorithm. 

\textbf{Comparison}: At the current stage of development, there is no corresponding baseline and sharable experimental protocol besides random sampling which has been proven to be a strong baseline. This makes it difficult to compare NAS search algorithms. Instead of blindly stacking certain techniques to increase performance, it is critical to have more ablation experiments on which part of the NAS design leads to performance gains. To support reproducibility and comparison, NASbenches \cite{ying2019bench} provide pre-computed performance measures for a large number of NAS architectures.

\section*{Acknowledgment}
This material is based upon work supported by the National Science Foundation under Award No. OIA-1946391; partially funded by Gia Lam Urban Development and Investment Company Limited, Vingroup and supported by Vingroup Innovation Foundation (VINIF) under project code VINIF.2019.DA19.

\section*{Disclaimer}
Any opinions, findings, and conclusions or recommendations expressed in this material are those of the author(s) and do not necessarily reflect the views of the National Science Foundation.



\Backmatter
\printbibliography
\end{document}